\documentclass[a4paper,11pt]{article}

\usepackage{graphicx}  
\usepackage{url}       
\usepackage{times}
\usepackage{natbib}
\usepackage[margin=25mm]{geometry}
\usepackage{amsmath}
\usepackage{graphicx,tabulary,multicol,multirow,booktabs,array,xcolor,wrapfig,cuted}
\usepackage{pbox}

\usepackage{linguex}

\usepackage[lofdepth,lotdepth]{subfig}
\usepackage{tabularx}

\usepackage[draft,inline,nomargin,index]{fixme}
\fxsetup{theme=color,mode=multiuser}
\FXRegisterAuthor{es}{esn}{\color{red}ES}
\FXRegisterAuthor{ec}{ecn}{\color{blue}EC}
\FXRegisterAuthor{al}{aln}{\color{green}AL}
\FXRegisterAuthor{pb}{pbn}{\color{black}PB}

\usepackage{linguex}

\usepackage{makeidx}
\makeindex

\title{Is Structure Necessary for Modeling Argument Expectations \\ in Distributional Semantics?}
\date{}

\author{Emmanuele Chersoni\\
       Aix-Marseille University\\
       \texttt{emmanuelechersoni@gmail.com}
  \and Enrico Santus\\
       Singapore University of Technology \\ and Design\\
       \texttt{esantus@mit.edu}
   \and Philippe Blache\\
       Aix-Marseille University\\
       \texttt{philippe.blache@univ-amu.fr}
   \and Alessandro Lenci\\
       University of Pisa\\
       \texttt{alessandro.lenci@unipi.it}
}

\begin{document}

\maketitle
\thispagestyle{empty}
\pagestyle{empty}

\begin{abstract}
Despite the number of NLP studies dedicated to thematic fit estimation, little attention has been paid to the related task of composing and updating verb argument expectations. The few exceptions have mostly modeled this phenomenon with \textit{structured} distributional models, implicitly assuming a similarly \textit{structured} representation of events.
Recent experimental evidence, however, suggests that human processing system could also exploit an unstructured ``bag-of-arguments" type of event representation to predict upcoming input. In this paper, we re-implement a traditional structured model and adapt it to compare  the
different hypotheses concerning the degree of structure in our event knowledge, evaluating their relative performance in the task of the argument expectations update. \\ \\
\end{abstract}

\section{Introduction}

An important trend of current research in linguistics and cognitive science aims at investigating the mechanisms behind anticipatory processing in natural language \citep{Kamide2003133,delong2005probabilistic, federmeier2007thinking,van2012prediction,willems2015prediction}. It is, indeed, uncontroversial that our cognitive system tries to predict incoming input on the basis of prior information, and this strategy is probably crucial for dealing with the rapidity of linguistic interactions \citep{christiansen2016now}. 
By means of different experimental paradigms, several studies have focused on the role of event knowledge in the activation of expectations on verb arguments \citep{McRae1998ModelingTI,McRae2005,Hare2009ActivatingEK,bicknell2010effects}. During sentence processing, verbs (\textit{arrest}) activate expectations on their typical argument nouns  (\textit{crook}), and nouns do the same for other arguments frequently co-occurring in the same events (\textit{cop-crook}). The explanation proposed by these studies is that the human ability to anticipate the incoming input depends on general knowledge about events and their typical participants. This knowledge, stored in the semantic memory, `reacts' to the linguistic stimulus: the more the processed information is coherent with a prototypical event scenario, the easier for the comprehension system is to constrain the range of the events potentially described by the sentence and to predict the upcoming sentence arguments.

According to one of the most influential accounts of event-based prediction, the event representation includes both the thematic roles and the lexical meanings of the arguments, 
as well as the relations between different roles. Therefore, 
there is the assumption of a \textit{structural distinction} between the participants filling the roles \citep{kim2016prediction} (i.e. some arguments are good \textit{agents}, other are good \textit{patients} etc.).
Such an account has been challenged by the experimental evidence for a `bag-of-arguments' mechanism of verb predictions, discussed by \citet{chow2016bag}. Their experiments with Event-Related Potentials (ERPs) focused on the N400 component,\footnote{The N400, one of the most well-studied ERP components, is a negative-going deflection that peaks around 400 ms after the presentation of a stimulus word.} whose amplitude is generally interpreted as reflecting the predictability of a word in context \citep{kutas1984brain}. One of their findings was that there is no significant difference in the N400 amplitude at the target verb in sentences like \ref{1a} and \ref{1b} (normal vs. \textit{role-reversed} argument configuration):

\ex.
\a. \label{1a} The restaurant owner forgot which \textbf{customer} the \textbf{waitress} had \textit{served} during dinner yesterday.
\b. \label{1b} The restaurant owner forgot which \textbf{waitress} the \textbf{customer} had \textit{served} during dinner yesterday.

That is to say, even if different roles are assigned to \textit{customer} and \textit{waitress} in \ref{1a} and \ref{1b}, this difference seems to have no impact on the N400 amplitude.

Given the lack of influence of the structural roles, in order to circumscribe their hypothesis (which we will henceforth refer to as the \textit{bag-of-arguments hypothesis}), the authors set up another experiment, in which they tested whether the predictions could be influenced also by other co-occurring words (i.e., not necessarily arguments; we will refer to this possibility as the \textit{bag-of-words hypothesis}).
In order to carry out this test, they compared the amplitudes for sentences like \ref{2a} and \ref{2b} (argument substitution), finding a significantly smaller N400 component for the first sentence type.

\ex.
\a. \label{2a} The \underline{exterminator} inquired which \textbf{neighbor} the \textbf{landlord} had \textit{evicted} from the apartment complex.
\b. \label{2b} The \underline{neighbor} inquired which \textbf{exterminator} the \textbf{landlord} had \textit{evicted} from the apartment complex.

\cite{chow2016bag} concluded that only the event \textit{arguments} can influence predictions about a verb, and that arguments are represented in a sort of unstructured collection (i.e., \textit{bag-of-arguments}). Therefore, according to them, predictions would be sensitive to the meaning of the arguments, but not to their structural roles, which are computed later. For example, the difference between typical agents and typical patients, according to this account, would not be included in the representation of an event.


In the last few years, a related issue has been debated in the field of distributional semantics, i.e. whether there is any added value in using structured representations of linguistic contexts over bag-of-words ones (e.g., contexts represented as co-occurrence windows). While structured models have been shown to outperform the latter in a number of semantic tasks \citep{Pado:Lapata:2007,Baroni:2010,Levy:Goldberg:2014}, some bag-of-word models proved to be extremely competitive, at least under certain parameter settings \citep{baroni2014don}. A recent paper by \citet{lapesa2017large} explicitly addressed the question of whether using structured distributional semantic models is worth the effort, by comparing the performance of syntax-based and window-based distributional models on four different tasks. The authors showed that, even after extensive parameter tuning, the former have a significant advantage only in one task out of four (i.e., noun clustering). Interestingly, in the discussion they leave
open the question of whether their results can generalize to linguistically challenging task such as the prediction of thematic fit ratings.


In this paper, we specifically investigate this point. The main questions we want to address are: what are the implications of the \textit{bag-of-arguments hypothesis} for current models of thematic fit? More precisely, is it really necessary to have \textit{structured} representations to carry out a thematic fit-related task, such as the argument expectation update?


In order to answer our questions, we implemented three models of argument expectations, adapting them to the above-mentioned hypotheses (i.e., \textit{structured} and \textit{unstructured}, the latter including both the \textit{bag-of-arguments} and the \textit{bag-of-words} hypothesis), and we compared their performance in
a binary selection task.

\section{Related Work}

One of the most influential distributional model of thematic fit was introduced by \citet{Baroni:2010}, who represented verb semantic roles with a \textit{prototype vector} obtained by averaging the dependency-based vectors of the words typically filling those roles (i.e. the \textit{typical fillers}). 
Within the Distributional Memory (DM) framework, which was based on syntactic dependencies, Baroni and Lenci used grammatical functions such as subject and object to approximate the thematic roles of agent and patient, and they measured role typicality by means of a Local Mutual Information score \citep{Evert2004TheSO} computed between verb, arguments and syntactic relations. The basic assumption is that the higher the distributional similarity of a candidate argument with a role prototype, the higher its predictability as a filler for that role will be. As a gold standard, the authors used the human-elicited thematic fit ratings collected by \citet{McRae1998ModelingTI} and \citet{pado2007integration}, and they evaluated the performance by measuring the correlation between these ratings and the scores generated by the model (as already proposed by \citet{Erk2010}). 

\citet{Lenci_composingand} later extended this `structured-approach' to account for the dynamic update of the expectations on an argument, which depends on how other roles in the sentences are filled. For instance, given the agent \emph{butcher} the expected patient of the verb \emph{cut} is likely to be \emph{meat}, while given the agent \emph{coiffeur} the expected patient of the same verb is likely to be \emph{hair}. By means of the same DM tensor, this study tested an additive and a multiplicative model \citep{mitchell2010composition} to compose the distributional information coming from the agent and from the predicate of an agent-verb-patient triple (e.g., \emph{butcher}--\emph{cut}--\emph{meat}), generating 
a prototype vector which represents the expectations on the patient filler, given the agent filler. The triples of the Bicknell dataset \citep{bicknell2010effects}, which were used for the first time to evaluate such a model, are still today, at the best of our knowledge, the only existing gold standard for this type of task.

Although the `structured-approach' to thematic fit was influential for a number of other works \citep{Sayeed2014combining,Sayeed2015AnEO,Greenberg2015ImprovingUV,Sayeed2016ThematicFE,santus2017measuring}, the task of modeling the update of the argument expectations has received relatively little attention. An exception is the work by \citet{Tilk2016EventPM}, who trained a neural network on a role-labeled corpus in order to optimize the distributional representation for thematic fit estimation. Their model was also tested on the task of the composition and update of argument expectations, where it was able to achieve a performance comparable to \citet{Lenci_composingand} on the triples of the Bicknell dataset.\footnote{\citet{Chersoni2016towards} presented a research work testing a similar method on the Bicknell dataset. However, their model does not really update argument expectations on the basis of other arguments, computing instead a global score of semantic coherence for the entire event representation, on the basis of the mutual typicality between all the participants.}
Notice that both the models of \citet{Lenci_composingand} and \citet{Tilk2016EventPM} necessarily rely on the hypothesis that the arguments are structurally distinct, since they are trained either on argument tuples containing fine-grained dependency information, or on sentences labeled with semantic roles.

Outside the specific area of study of thematic fit modeling, \citet{ettinger2016modeling} successfully used a type of unstructured representation for another sentence processing-related task, i.e. modeling N400 amplitudes with distributional spaces. The authors proposed a method based on word2vec \citep{mikolov2013efficient} to build vector representations of sentence context, and to quantify the relation of an upcoming target word to the context. After training their word vectors on the Ukwac corpus \citep{Baroni2009} with the Skip-Gram architecture, they modeled the mental state of a comprehender at a certain point of a sentence as the average of the vectors of the words in the sentence up to that point. The predictability of a target word in a sentence was measured as the cosine similarity between its vector, and the context-vector obtained by averaging the vectors of the preceding words.  
Ettinger and colleagues tested their method on the sentences used in the ERP study by \citet{federmeier1999rose}, in which three different conditions were defined, and they observed that the context-target similarity scores across conditions were following the same pattern of the N400 amplitudes of the original experiment. 
Thus, this work shows how data on N400 variations can be modeled even by means of vectors with minimal or no syntactic information. 

\section{Experiments}

\noindent{}\textbf{Rationale}. \citet{Baroni:2010} computed the thematic fit for a candidate $filler$ (e.g., \textit{policeman}) in an argument $slot$ (e.g., agent) of an $input$ lexical item  (e.g., \textit{arrest}) as the similarity score between the vector of the candidate $filler$ and a prototype of the typical $slot$ filler, built by summing the vectors for the top-\textit{k} most typical fillers of $input$ for $slot$ (e.g., the typical agents for the \textit{arrest}-event, such as \textit{cop}, \textit{officer}, \textit{policewoman}, etc.). In this model, syntactic relations were used to approximate verb-specific semantic roles and to identify the most typical fillers. For example, the agent role is approximated by the subject relation, so that the typical $fillers$ for the agent $slot$ are the typical subjects of the $input$ verb. Similarly, the patient role is approximated by the object relation, so that the typical $fillers$ for the patient $slot$ are the typical objects of the $input$ verb.


We propose an extension of the model by \citet{Lenci_composingand} and we interpret thematic fit as \textit{the expectation of an argument} (i.e., what the prototype vector is meant to represent: $EX_{slot}(input)$), claiming that the update on expectations for a filler caused by new input (e.g. a verb combining with an agent) could be modeled by means of a function $f(x)$ that combines the prototypes built for every input:

\begin{equation} \label{expectations} 
EX_{slot}(\langle input_1, input_2 \rangle) = 
f(EX_{slot_1}(input_1), EX_{slot_2}(input_2))
\end{equation}

\noindent{}where the function $f(x)$ is the sum or the pointwise multiplication between the prototype vectors. Once the expectations are calculated, the $filler$ fit for the $slot$ of $\langle input1, input2 \rangle$ can be computed by measuring the similarity (e.g., by vector cosine) between the $filler$ and the expectations. As an example, if we want to estimate how likely is \textit{burglar} as a patient of \textit{the policeman arrested the...}, 
we build a prototype out of the vectors of typical objects co-occurring with the subject \textit{policeman-n}, 
then we do the same for the vectors of typical objects of the verb \textit{arrest-v}, 
and finally we combine the prototype vectors through $f(x)$, by either sum or multiplication. At this point, we can estimate the $filler$ fit by calculating the following similarity:


\begin{multline} \label{exampleExpectations} 
EX_{patient}(burglar | \langle police, arrest \rangle) =  \\
sim(burglar, f(EX_{cooc\_patient}(policeman), EX_{patient}(arrest)))
\end{multline}

\noindent{}Since distributional similarity is used as a measure of the predictability of a filler for a certain role, we expect that the thematic fit score of \textit{burglar} for the patient $slot$ of $\langle policeman, arrested \rangle$ 
will be much higher than for \textit{singer-n}. Indeed, \textit{burglars} are more typical patients in this type of situation than \textit{singers} are. Notice that while in \citet{Lenci_composingand} the update function modified the association scores between the predicate and the fillers, in the present case $f(x)$ directly composes the prototype vectors associated with $\langle input1, input2 \rangle$.\\

\begin{table}[ht]
\small
\centering
\begin{tabularx}{\textwidth}{|l|l|X|}
\hline
              & \textbf{Target}  & \textbf{Fillers}                                                                                                                                                                                             \\ \hline
\textbf{BOW}  & steal-v           & car-n, money-n, show-n, base-n, thief-n, good-n, item-n, property-n, someone-n, horse-n, limelight-n, vehicle-n, attempt-n, cattle-n, food-n, wallet-n, bike-n, identity-n, thunder-n, key-n             \\ \hline
\textbf{BOA}  & steal-v           & show-n, money-n, car-n, base-n, food-n, thunder-n, march-n, limelight-n, horse-n, idea-n, key-n, wallet-n, heart-n, property-n, jewel-n, identity-n, cattle-n, body-n, purse-n, treasure-n      \\ \hline
\textbf{DEPS} & steal-v (agent)   & thief-n, someone-n, man-n, burglar-n, gang-n, money-n, robber-n, handbag-n, criminal-n, wallet-n, computer-n, thou-n, horse-n, equipment-n, boy-n, crook-n, disciple-n, cash-n, somebody-n, dog-n                    \\ \hline
\textbf{DEPS} & steal-v (patient) & show-n, money-n, car-n, base-n, food-n, thunder-n, march-n, limelight-n, horse-n, idea-n, key-n, wallet-n, heart-n, property-n, identity-n, cattle-n, jewel-n, body-n, purse-n, treasure-n \\ \hline
\end{tabularx}
\caption{Top-20 filler nouns for the word \textit{steal-v} in our three models (for DEPS, we provide the fillers for the agent and the patient slot, while in the other models there is no distinction).}
\label{fillers}
\end{table}

\noindent{}\textbf{Models}. In our experiments, we compared three different distributional semantic models (henceforth DSMs), all inspired by \citet{Lenci_composingand}: i) a structured model, which is similar to the one presented in \citet{Lenci_composingand} (\textbf{DEPS}); ii) a variation of this system, modeling the \textit{bag-of-arguments hypothesis} (\textbf{BOA}); iii) a baseline relying on the \textit{bag-of-words hypothesis} (\textbf{BOW}). The key difference between our models is to be found in the selection of the fillers (see Table \ref{fillers}):
\begin{itemize}
\item \textbf{DEPS}: Similarly to Lenci's system, DEPS makes use of information on specific syntactic relations to select role fillers: the agent-role prototypes will be built out of the most typical subjects, the patient-role prototypes out of the most typical objects, and so on (see the last two rows of Table \ref{fillers}).\footnote{Since the roles are approximated by syntactic relations identified by the parser (i.e. Malt-parser \citep{Nivre05maltparser:a}), their accuracy is subordinate to the accuracy of the parser. Ideally, we would expect clean lists of fillers for the typical subjects (agents) and objects (patients) of a verb, but -- as it can be seen in Table \ref{fillers} -- this is not the case.} This means that not only the semantic information is taken in consideration (e.g. \textit{policeman}), but also the thematic role of the filler (e.g. \textit{sbj:policeman}, \textit{obj:policeman}, etc.). Since dependencies are used to filter fillers entering in the role representation, this model is the closest one to theories assuming \textit{structured} event knowledge. 

\item \textbf{BOA}: Almost identical to the DEPS model, except for the fact that the most typical arguments are not bound to a specific syntactic slot. Indeed, according to \citet{chow2016bag}, the arguments of a verb like \textit{to serve} (\textit{customer}, \textit{waitress}, \textit{tray}, etc.) are represented like an \textit{unstructured} collection. In this type of model, thus, the top-\textit{k} typical fillers will include \textit{all the strongly associated arguments}, abstracting away from the specific syntactic relation.


\item\textbf{BOW}: In this baseline, the typical fillers are not arguments, but words typically co-occurring with the targets in a window of fixed width (possibly having no syntactic relation to the targets).
\end{itemize}

Going back to the previous example, the core idea of the DEPS model is that processing a sentence fragment like \textit{the policeman arrested the...} leads to the activation only of the typical \textit{patients} of such events, since the event knowledge is assumed to be structured. Therefore, the predictability of an argument is measured in terms of its similarity with the prototype built out of the activated patients.

On the other side, the BOA model assumes no distinction between the arguments (i.e., whether they are agents, patients, locations or others), and consequently the sentence fragment above would activate all the typical arguments of the verb \textit{arrest}. This means that the predictability of an argument will be equivalent to its similarity with the prototype of a generic argument of the verb.

Finally, the BOW baseline has no notion of structure at all, not even the underspecified argument relation of the BOA model, and thus the prototypes of this model are just representations of the typical neighbors of the target words. It should be recalled at this point that a \textit{bag-of-words} account of prediction was ruled out by the experimental results by \citet{chow2016bag}, since only arguments turned out to have an impact. Nonetheless, since we have chosen a \textit{bag-of-words} model with a very narrow window (i.e., two words on the left and right of the target), BOW could also capture indirectly syntactic information (i.e., words frequently co-occurring with the targets within a narrow window are very likely to be also syntactically related to them). Therefore, we expect it to be a reasonably strong baseline.\\

\noindent{}\textbf{Corpus and DSMs}. 
Distributional information is derived from the concatenation of the British National Corpus \citep{leech1992100} and of the Wacky \citep{Baroni2009} corpus. Both were parsed with the Maltparser \citep{Nivre05maltparser:a}. 
From this concatenation, we built a dependency-based DSMs, where the tuples are weighted by means of Positive Local Mutual Information (PLMI, \citet{Evert2004TheSO}). 
Given the co-occurrence count $O_{trf}$ of the target $t$, the syntactic relation $r$ and the filler $f$, we computed the expected count $E_{trf}$ (i.e., the simple joint probability of indipendent variables, corresponding to the product of the probabilities of the single events).\footnote{The DSM were built by means of the scripts of the DISSECT framework \citep{dinu2013dissect}}

The PLMI for each target-relation-filler tuple is computed as follows:

\begin{equation} \small \label{1}
LMI(t,r,f) = log\left ( \frac{O_{trf}}{E_{trf}}\right ) * O_{trf}
\end{equation}

\begin{equation} \small \label{2}
PLMI(t,r,f) = max(LMI(t,r,f), 0)
\end{equation}

Our DSM contains 28,817 targets (i.e., all nouns and verbs with frequency above 1000 in the training corpora), and all syntactic relations were included.\footnote{We added the extra relation VERB, accounting for the link between typically co-occurring subjects and objects. An analogous relation was already in \citet{Baroni:2010}.} We also built a window-based DSM to extract co-occurrence information for the BOW model, counting only the co-occurrences between the nouns and the verbs of the list above within a word window of width 2.\\


\noindent{}\textbf{Prototypes} The prototypes of all models were built out of the vectors of the $k$ most typical fillers for each model type, and we tested 10, 20, 30, 40, and 50 as values of $k$.\footnote{The choice of the parameter range is in line with previous NLP studies on thematic fit \citep{Sayeed2015AnEO,Greenberg2015ImprovingUV}.} 

As in previous studies, PLMI values were used as typicality scores: in the DEPS model, the typicality ranking of the fillers for a given role takes into account \textit{only the fillers occurring in the corresponding syntactic slot} (e.g. the subject for the agent, the object for the patient etc.), whereas in the BOA model the typicality of a filler only depends on the PLMI score with the target, thus ignoring the type of syntactic relation.\footnote{It goes without saying that using syntactic functions to identify the fillers of semantic roles is just an approximation. Nonetheless, the good performances reported by syntax-based thematic fit estimation systems suggest that, at least for agents and patients, such an approach is empirically justified.} As for the BOW baseline, the words used for building the prototype are simply co-occurring with the targets within a word window of width 2, and such co-occurrences have been PLMI-weighted as well.\\


\noindent{}\textbf{Compositional Functions}. The compositional functions that we used to combine the prototypes are the vector sum and the pointwise vector multiplication \citep{mitchell2010composition}. An important difference between the compositional functions lies in the fact that, while the sum retains the dimensions that are not shared by both prototype vectors, the multiplication sets them to zero those dimensions. This has an obvious impact on the computation of the cosine, as it could drastically reduce the number of dimensions on which the similarity score is computed.\\

\noindent{}\textbf{Datasets and Evaluation}.  The models were tested on the datasets from the ERP experiments by \citet{bicknell2010effects} and \citet{chow2016bag}.

The Bicknell dataset was introduced to test the hypothesis that the typicality of a verb direct object depends on the subject argument. With this purpose in mind, the authors selected 50 verbs, each paired with two agent nouns that significantly changed the scenario evoked by the subject-verb combination. They obtained typical patients for each agent-verb pair by means of production norms, and they used such data to generate triples where the patient was congruent with the agent and with the verb. For each congruent triple, an incongruent triple was generated as well, by combining each verb-congruent patient pair with the other agent noun, in order to have items describing atypical situations.

The final dataset is composed by 100 plausible-implausibile triples, which were used to build the sentences for a self-paced reading and for an ERP experiment. The subjects were presented with sentence pairs such as:

\begin{itemize}
\item \textit{The journalist checked the spelling of the last report.} (plausible)
\item \textit{The mechanic checked the spelling of the last report.} (implausible). 
\end{itemize}

\citet{bicknell2010effects} reported shorter reading times and smaller N400 amplitudes for the plausible condition. The goal, for a thematic fit model of the argument expectations update, is to assign a higher cosine similarity score to the plausible triple, as in \citet{Lenci_composingand}.
Moreover, \citet{Tilk2016EventPM} evaluated their systems on two different versions of this task, since the triple pairs can be created by combining either triples differing  only for the agent, or triples differing only for the patient. Following the terminology from this latter study, we will refer to \textbf{Accuracy 1} meaning the accuracy of the models in scoring differing-by-patient triples, and to \textbf{Accuracy 2} meaning the accuracy in the classification of the differing-by-agent ones.

We also turned into similar triples the 50 verb-arguments combinations of the role reversal experiment by \citet{chow2016bag}, by creating triple pairs corresponding to the normal and to the role-reversed condition. For example, the sentences in Example (1) were turned into the form: \textit{customer-n waitress-n serve-v} (normal) and \textit{ waitress-n customer-n serve-v} (role-reversed). Notice that we preserved the order in which the experimental subjects saw the arguments and the verb, with the latter at the end. Consequently, instead of composing the prototype vectors of the typical fillers of the patient role given an agent and a predicate, as we did for the Bicknell dataset, we derive the expectation vector for the verb from the composition of the prototypes of the typical predicates of the agent and of the patient. 

The binary selection task is the same used with the Bicknell dataset, the only difference being that the goal for our models is to assign higher scores to the triples in the standard argument configuration (i.e., the expectation vector should be closer to the verb vector in the normal condition). Only the DEPS results are reported for the Chow dataset, because unstructured models assign exactly the same score to normal and role-reversed triples (independently of the order in which the prototypes of the head verb for each argument are created, the combined prototype will be the same). This is, of course, consistent with the report of the ERP experiment by Chow and colleagues, who found no differences in the N400 amplitudes elicited by the two sentence types. 

The performance of the DEPS model on the Chow dataset is of particular interest, as the model has the structural information that is lacking in the other two. If DEPS has to reproduce the N400 pattern found by Chow and colleagues, the scores for the normal and for the role-reversed conditions should not differ significantly.

\section{Results}

In Table \ref{Accuracy1} and \ref{Accuracy2}, we report the results for the three model types on the Bicknell dataset for the two kinds of prototype composition and $k = 20$. This latter value is the most common in the literature \citep{Baroni:2010,Greenberg2015ImprovingUV}, and the one that gave us the highest accuracy scores.

The DEPS model is almost always the best performing one on the Bicknell dataset, with the exception of a single drop for the multiplicative model in the Accuracy 1 evaluation. The sum turned out to be the most efficient combination function in the majority of the models, and a possible explanation is that the application of multiplication to dependency-based prototype vectors led to sparsity problems. The results obtained by the DEPS Sum model are the highest ones, and the Accuracy 2 score for $k = 20$ is extremely close to the best performance reported in \citet{Lenci_composingand}
(73\%).\footnote{The only result available for comparison in the literature is the one obtained by \citet{Lenci_composingand} on the Accuracy 2 task, since the evaluation of \citet{Tilk2016EventPM} was carried out on a way smaller subset of the Bicknell dataset (64 triples).}
The task of classifying differing-by-patient triples turns out to be harder, as the accuracies are lower and none of the models is significantly better than a random baseline ($p$-values were computed with the $\chi^2$ statistical test)\footnote{Also the Accuracy 1 scores reported by \citet{Tilk2016EventPM} confirm the higher difficulty of this version of the task.}, whereas the Accuracy 2 scores of both the Deps Sum Model and the Deps Multiplication Model have a significant advantage (for both of them, $p < 0.05$).

\begin{figure*}[ht]
\begin{center}
\includegraphics[height = 6cm]{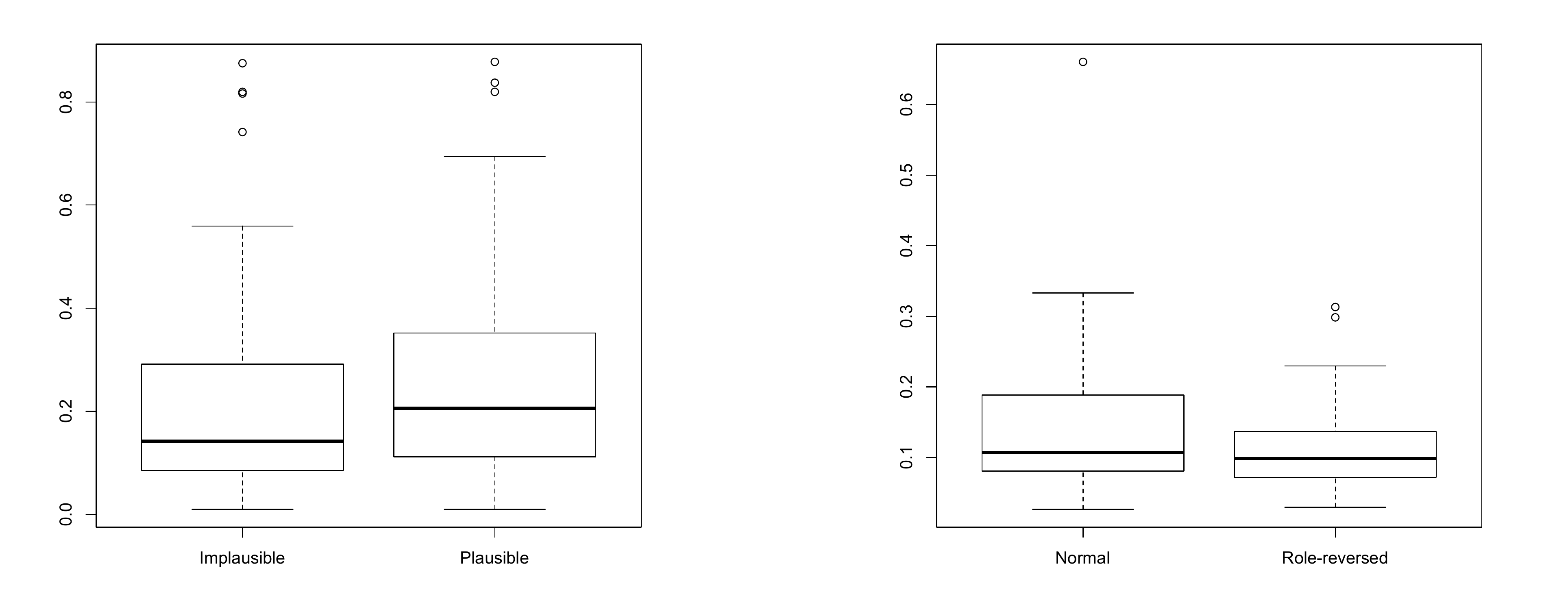}
\caption{Cosine scores assigned by DEPS-sum model ($k = 20$) for the Bicknell dataset (left) and for the Chow dataset (right).}
\label{boxplots}
\end{center}
\end{figure*}

We also carried out the Wilcoxon rank sum test on the scores generated by all models, and we found that the DEPS-sum model is the only one that manages to assign significantly different scores to the sentences in the two conditions ($W = 5660, p < 0.05$; for all the other models, $p > 0.1$) (see Figure \ref{boxplots}).
The BOA model was found instead to be worst performing one, even lower than the BOW baseline, and often the recorded accuracy scores are very close to a random baseline. Also, the differences between conditions were far from significance in any of the versions of the model.

\begin{table}[ht]
\small \centering
\begin{tabular}{|c|c|c|}
\hline
Model & Sum & Multiplication \\
\hline
BOW & 59\% & 57\% \\ 
\hline
BOA & 53\% & 59\% \\
\hline
DEPS & \textbf{62\%} & 56\% \\
\hline
\end{tabular}
\caption{Accuracy 1 on Bicknell (100\% coverage) for $k = 20$}
\label{Accuracy1}
\end{table}

\begin{table}[ht]
\small \centering
\begin{tabular}{|c|c|c|}
\hline
Model & Sum & Multiplication \\
\hline
BOW & 60\% & 56\% \\ 
\hline
BOA & 58\% & 57\% \\
\hline
DEPS & \textbf{72\%} & 68\% \\
\hline
\end{tabular}
\caption{Accuracy 2 on Bicknell (100\% coverage) for $k = 20$}
\label{Accuracy2}
\end{table}

Figure 2 shows the performance variation of the Sum models on Bicknell dataset, while varying the number of fillers used to build the prototype. At a glance, we can observe that DEPS models achieve higher accuracies with fewer fillers. This is kind of expected, since the good performances of such models are likely to be due to a more restrictive selection of the fillers. With higher values of $k$, the selection of more weakly-related fillers is probably introducing noise in the prototype.
On the other hand, BOA models slightly improve when more fillers are used, but in general their performance is almost always equivalent to BOW models. This indicates, in our view, that the underspecified dependency relation of the BOA model is insufficient to build a precise representation of the expectations on an upcoming argument, unless a larger number of fillers is taken into account. Moreover, even if typical arguments are selected by virtue of a dependency relation, the absence of information on the dependency type makes these models essentially equivalent to window-based ones. Finally, as for the difference between the scores in the two conditions, the Wilcoxon rank sum test returns a significant difference only for the DEPS Sum model with $k = 10, 20$ (in both cases, $p < 0.05$).

\begin{figure*}[ht]
\begin{center}
\includegraphics[height = 6cm]{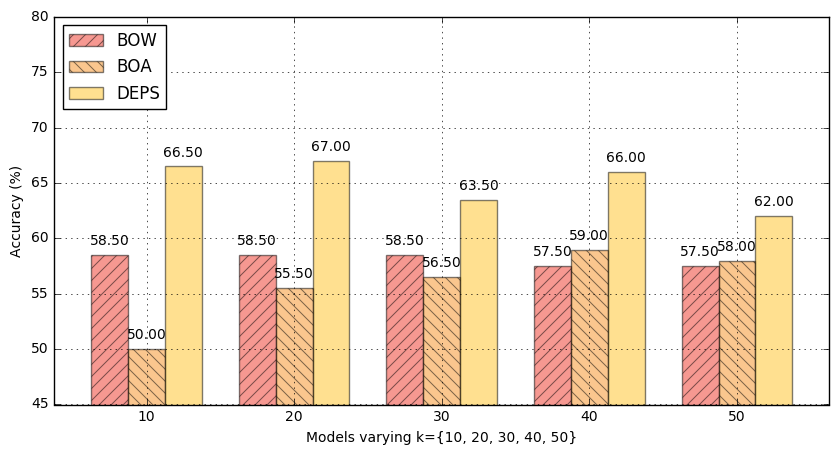}
\caption{Average of Accuracy 1 and Accuracy 2 on Bicknell dataset, for all the Sum models for different values of $k$}
\label{fig:accuracy}
\end{center}
\end{figure*}

Concerning the performance of DEPS on the Chow dataset, it can be seen in Table \ref{Accuracy3} that the system identifies the triple in the normal condition with a level of accuracy between 62\% and 68\%. The scores are quite steady, independently from $k$, and again, the Sum models are generally performing better (but never significantly better than a random baseline: for all parameter settings, $p > 0.05$). 

Interestingly, after applying the Wilcoxon rank sum test, it turns out that the differences between the assigned scores never differ significantly between the normal and the argument reversal condition (for all values of $k$, $p > 0.05$; see also Figure \ref{boxplots}). This result is coherent with the outcome of the role-reversal experiment by Chow and colleagues, who found no difference between the N400 elicited by the two sentence types. In other words, structural information does not help in predicting the upcoming verb.


\begin{table}[ht]
\small \centering
\begin{tabular}{|c|c|c|}
\hline
k & Sum & Multiplication \\
\hline
10 & 68\% & 66\% \\
\hline
20 & 62\% & 64\% \\
\hline
30 & 64\% & 62\% \\
\hline
40 & 64\% & 62\% \\
\hline
50 & 66\% & 62\% \\
\hline
\end{tabular}
\caption{Accuracy on Chow (100\% coverage) for the DEPS model for different values of $k$}
\label{Accuracy3}
\end{table}

In the same way as the unstructured representations used by \citet{ettinger2016modeling}, our models show that the distributional similarity between a target and its context (a structured one, in our case) can accurately reflect the N400 amplitude patterns found in the experimental studies. Notice however that only a model based on the notion of a structured event knowledge was able to mirror the patterns of both the studies of \citet{bicknell2010effects}.
Together with
the better performance reported on the Bicknell dataset for the binary classification task, these results suggest that the presence of structural information is an advantage for distributional models of thematic fit.

\section{Conclusions}

In this paper, we addressed the question of whether structured information is necessary to model the argument expectation update. With this purpose in mind, we have implemented a traditional system for composing and updating thematic fit estimations \citep{Lenci_composingand} and we adapted it to model both structured and unstructured representations (the latter including both the \textit{bag-of-arguments} and the \textit{bag-of-words} hypotheses). We compared the performance of these models on the binary selection task of the argument expectation update and on their ability to replicate the experimental results from the studies by \citet{bicknell2010effects} and \citet{chow2016bag}.

Our results show that structured models perform better in a task of composing and updating argument expectations, and can reproduce the ERP results reported for both datasets. On the other hand, the \textit{bag-of-arguments} model had lower scores in the classification task on the Bicknell dataset, and it was not able to discriminate between the plausible and the implausible condition.

It should be recalled that the \textit{bag-of-arguments} hypothesis was proposed to account for the results of an experiment on initial verb predictions, where the participants could see the verb only at the end (see Examples 1 and 2). In absence of any cue facilitating the mapping between arguments and syntactic positions (consider also that the arguments in the dataset do not differ by animacy), it is reasonable to hypothesize a delay in the assignment of the thematic roles. Moreover, as already pointed out by \citet{kim2016prediction}, the N400 component is generally not sensitive to the implausibility derived by thematic role reassignments, but the presence of event knowledge violation in such cases can be signaled by other ERP components.\footnote{\citet{kim2016prediction} point out that role-reversed sentences typically elicit en enhanced P600 components compared to their plausible counterparts (see also the experiments in \citet{kim2005independence} and \citet{kim2011conflict}).} 
In sum, the idea of a structure in the event knowledge does not seem to be incompatible with the findings of Chow and colleagues, since our structured DSMs replicated the lack of significant differences between normal and role-reversed sentences. On the other side, models with no structural information struggle in modeling the results of datasets where the items differ for their context-sensitive argument typicality, like the one from \citet{bicknell2010effects}.\footnote{The `bag-of-arguments' mechanism described by \citet{chow2016bag} actually concerns a very early stage of the comprehension process. Moreover, in a later response article, \citet{chow2016prediction} brought evidence that verb predictions become sensitive to structural roles of the arguments if more time is available for prediction. \\ We thank one of our reviewers for pointing this out.} 

The performance of the DEPS model also complies with the conclusions of \citet{ettinger2016modeling}, which showed how DSMs could be used to reproduce the N400 variations. Such a component is known to be tied to the general semantic relatedness of a target word to its sentential context, and not to syntactic anomalies.\footnote{The study of \citet{chow2016bag} is an example of experimental evidence for this claim, but see also the results from \citet{fischler1985brain} on the N400 insensitivity to the introduction of negations.} From this point of view, it is interesting that our structured models, despite their coherence with the ERP results by \citet{chow2016bag}, are still able to distinguish the sentence in the normal condition from the role-reversed one with an accuracy always above 60\%. Future research could explore in which measure thematic fit models can be sensitive to differences between syntactically-composed representation.
Finally, with reference to \citet{lapesa2017large}, our results make the expectation update task a good candidate for being among those that clearly benefit from using fine-grained syntactic information, as it seems to require knowledge about the relation types and about the interdependencies between participants. 

Future works might aim at comparing these model types on other NLP tasks, to check how many of them effectively take advantage from structured representations. For the moment, we can conclude that structure is an important added value for thematic fit models.

\section*{Acknowledgments}

This work has been carried out thanks to the support of the A*MIDEX grant (n°ANR-11-IDEX-0001-02) funded by the French Government ``Investissements d'Avenir" program. \\
We would like to thank the anonymous reviewers for their comments and for their helpful suggestions.

\bibliographystyle{chicago}
\bibliography{emnlp2017}

\end{document}